\documentclass[sn-mathphys-num, iicol]{sn-jnl}% Math and Physical Sciences Numbered Reference Style 
\usepackage{graphicx}%
\usepackage{multirow}%
\usepackage{amsmath,amssymb,amsfonts}%
\usepackage{amsthm}%
\usepackage{mathrsfs}%
\usepackage[title]{appendix}%
\usepackage[table,xcdraw,dvipsnames]{xcolor}%
\usepackage{textcomp}%
\usepackage{manyfoot}%
\usepackage{booktabs}%
\usepackage{algorithm}%
\usepackage{algorithmicx}%
\usepackage{algpseudocode}%
\usepackage{listings}%
\usepackage{tabularx}
\usepackage{array}
\usepackage{url}
\newcolumntype{L}{>{\raggedright\arraybackslash}X}
\usepackage{subcaption}
\usepackage[skip=0.333\baselineskip]{caption}
\usepackage[capitalise]{cleveref}
\usepackage{lmodern}
\usepackage{anyfontsize}

\begin{document}

\title[Article Title]{FLAASH: Flow-Attention Adaptive Semantic Hierarchical Fusion for Multi-Modal Tobacco Content Analysis}


\author*[1]{\fnm{Naga VS Raviteja} \sur{Chappa}}\email{nchappa@uark.edu}

\author[2]{\fnm{Page Daniel} \sur{Dobbs}}\email{pdobbs@uark.edu}

\author[3]{\fnm{Bhiksha} \sur{Raj}}\email{bhiksha@cs.cmu.edu}

\author[1]{\fnm{Khoa} \sur{Luu}}\email{khoaluu@uark.edu}

\affil*[1]{\orgdiv{Department of EECS}, \orgname{University of Arkansas}, \orgaddress{ \city{Fayetteville} \state{AR}, \country{USA}}}

\affil[2]{\orgdiv{Center for Public Health and Technology}, \orgname{University of Arkansas}, \orgaddress{ \city{Fayetteville} \state{AR}, \country{USA}}}

\affil[3]{\orgdiv{School of Computer Science}, \orgname{Carnegie Mellon University}, \orgaddress{ \city{Pittsburgh} \state{PA}, \country{USA}}}


\abstract{
The proliferation of tobacco-related content on social media platforms poses significant challenges for public health monitoring and intervention. This paper introduces a novel multi-modal deep learning framework named Flow-Attention Adaptive Semantic Hierarchical Fusion (FLAASH) designed to analyze tobacco-related video content comprehensively. FLAASH addresses the complexities of integrating visual and textual information in short-form videos by leveraging a hierarchical fusion mechanism inspired by flow network theory. Our approach incorporates three key innovations, including a flow-attention mechanism that captures nuanced interactions between visual and textual modalities, an adaptive weighting scheme that balances the contribution of different hierarchical levels, and a gating mechanism that selectively emphasizes relevant features. This multi-faceted approach enables FLAASH to effectively process and analyze diverse tobacco-related content, from product showcases to usage scenarios. We evaluate FLAASH on the Multimodal Tobacco Content Analysis Dataset (MTCAD), a large-scale collection of tobacco-related videos from popular social media platforms. Our results demonstrate significant improvements over existing methods, outperforming state-of-the-art approaches in classification accuracy, F1 score, and temporal consistency. The proposed method also shows strong generalization capabilities when tested on standard video question-answering datasets, surpassing current models. This work contributes to the intersection of public health and artificial intelligence, offering an effective tool for analyzing tobacco promotion in digital media.}


\maketitle

\section{Introduction}

The proliferation of short-form video content on social media platforms has transformed the landscape of public health communication, particularly regarding critical issues like tobacco use and promotion. This shift presents unique challenges in content analysis, as traditional methods often fail to capture the intricate interplay of visual and textual elements characteristic of these platforms.

Recent advancements in Large Language Models (LLMs) have shown remarkable capabilities in text processing~\cite{liu2024visual, zhu2023minigpt, li2022blip, li2023blip, dai2024instructblip}. However, their applications to video content, especially in the context of tobacco-related material, still need to be improved. The inadvertent promotion of tobacco products on social media has raised significant concerns among researchers and health advocates~\cite{lee2023cigarette, kwon2020perceptions, jancey2023promotion, sun2023vaping}, highlighting the need for more sophisticated analytical tools.

\begin{figure}
    \centering
    \includegraphics[width=\linewidth]{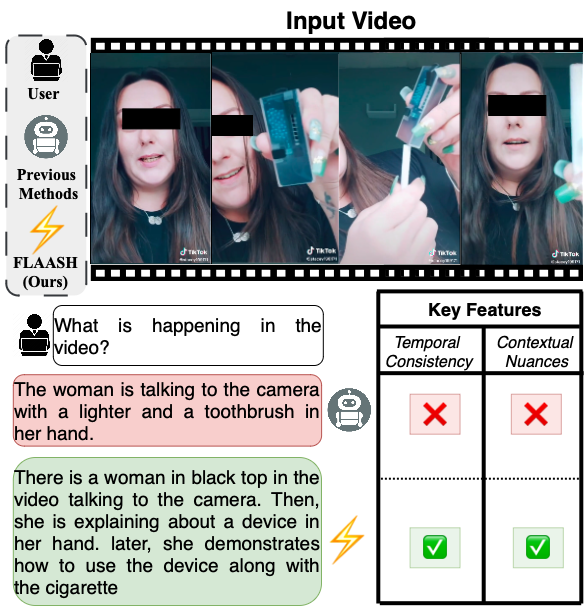}
    \caption{Comparison of FLAASH with previous methods~\cite{maaz2023video, tan2024koala} in analyzing tobacco-related video content. The input video shows a woman demonstrating a tobacco product. Our FLAASH model provides a more detailed and accurate scene description, capturing temporal consistency and contextual nuances. The key features highlighted show FLAASH's superior performance in these aspects compared to previous methods. \textbf{Best viewed in color and zoom.}}
    \label{fig:motivation}
\end{figure}

As illustrated in~\cref{fig:motivation}, previous methods struggle with several critical aspects of tobacco-related content analysis:

\textbf{Modality Imbalance}: Existing approaches often fail to adequately balance video and textual features, leading to suboptimal fusion and incomplete understanding of the content.
\textbf{Temporal Dynamics}: The crucial temporal aspects of tobacco use patterns and product demonstrations in videos are frequently overlooked or poorly captured.
\textbf{Contextual Nuances}: Subtle contextual cues present in both visual and textual modalities are often missed, resulting in less accurate interpretations of tobacco-related content.
\textbf{Content Diversity}: The wide variety of tobacco-related content, from close-up product displays to wide-angle social scenarios, poses significant challenges for current models in understanding and analyzing diverse visual contexts.

\begin{figure*}
    \centering
    \includegraphics[width=\linewidth]{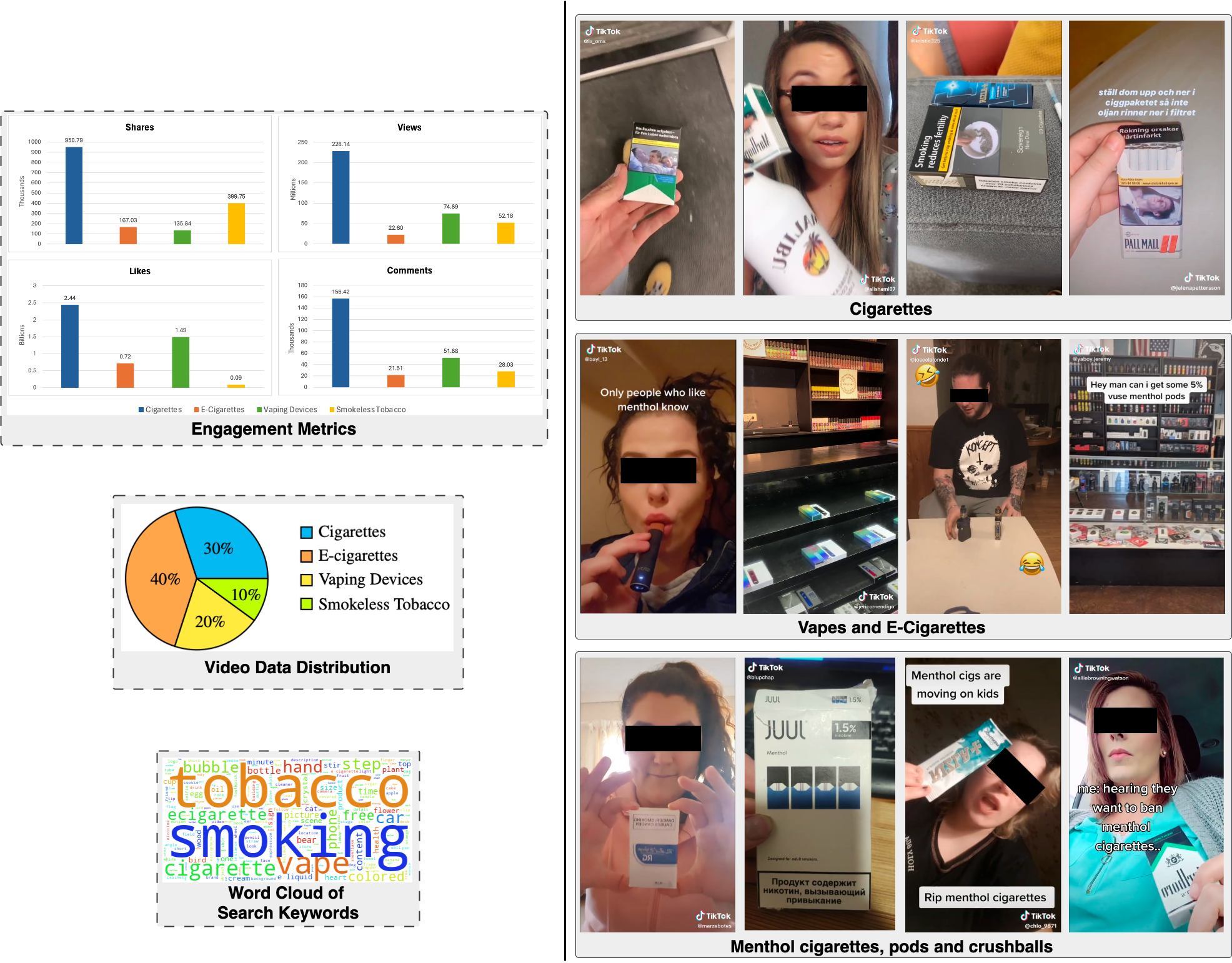}
    \caption{\textbf{(Left)} We present different metrics and a word cloud from our MTCAD dataset. \textbf{(Right)} Sample images from the MTCAD dataset categorized by tobacco product type. \textbf{Best viewed in color and zoom.}}
    \label{fig:dataset_samples}
\end{figure*}

~\Cref{fig:dataset_samples} further illustrates the diversity of tobacco products in our MTCAD dataset, underscoring the complexity of the content analysis task and the limitations of existing methods in handling such varied visual information.

To address these challenges, we introduce a novel multi-modal analysis framework named Flow-Attention Adaptive Semantic Hierarchical Fusion (FLAASH),  specifically designed for tobacco-related content in social media videos. The key contributions of this work can be summarized as follows:

\begin{itemize}
    \item \textbf{Flow-Attention Adaptive Hierarchical Cross-Attention Fusion (FlowAHCAF)}: We propose a revolutionary multi-modal fusion mechanism that leverages principles from flow network theory. FlowAHCAF incorporates:
    \begin{itemize}
        \item A hierarchical structure with multiple abstraction levels, capturing fine-grained details and high-level concepts.
        \item An adaptive weighting scheme that dynamically adjusts the importance of different hierarchical levels.
        \item A gating mechanism that selectively emphasizes relevant features at each level.
    \end{itemize}
    This approach enables nuanced, multi-scale interactions between visual and textual features, offering a more comprehensive analysis of tobacco-related content across various presentational contexts.
    
    \item \textbf{Composite Multi-Task Learning Objective}: We develop a sophisticated loss function that addresses product classification, sentiment analysis, temporal coherence, and feature discrimination. This multi-faceted approach ensures a holistic understanding of tobacco-related content, capturing both explicit and implicit aspects of usage and promotion.
    
    \item \textbf{Improved Performance and Insights}: We demonstrate substantial improvements over existing methods in critical analytical tasks on a diverse tobacco-related social media content dataset. Additionally, we provide insights into the model's decision-making process, offering a deeper understanding of multi-modal information processing in this context.
    
    \item \textbf{Public Health Impact}: By bridging the gap between cutting-edge machine learning techniques and public health advocacy, our work provides valuable tools for researchers and policymakers, paving the way for more effective, data-driven strategies in addressing tobacco use and promotion on social media platforms.
\end{itemize}

This paper explores our methodology in-depth, followed by extensive experimental results and analysis. The proposed approach not only advances the field of multi-modal content analysis but also opens new avenues for analyzing complex, multi-modal content in public health contexts, with potential applications extending beyond tobacco-related issues to other areas of public health and social media analysis.

\vspace{-3mm}
\section{Related Work}

\noindent \textbf{Multimodal Research with Large Language Models:} The field of video analysis has seen significant progress through deep learning techniques~\cite{wang2021actionclip, chappa2023sogar, tang2018vctree, zhong2024learning, zareian2020bridging, zareian2020learning, lu2016visual, zhong2021learning, ye2021linguistic, nguyen2024type,nguyen2023hig, chappa2023spartan, chappa2024flaash, truong2022otadapt, jalata2022eqadap, chappa2020squeeze, chappa2024advanced, chappa2024public}. However, these approaches often lack the crucial element of language comprehension, limiting their ability to fully interpret complex video content. Recognizing this limitation, researchers have begun to explore the integration of textual information~\cite{chappa2024react} to enhance the overall understanding capabilities of video analysis systems.

The emergence of powerful language models~\cite{gpt3,instructgpt,chatgpt} has sparked a new wave of research in multimodal learning. This has led to the development of sophisticated models that can process both visual and textual information simultaneously. These advanced methods~\cite{gpt4v, qwen_vl, otter, gemini,deepspeed_visualchat} have demonstrated remarkable capabilities in tasks such as visual-textual dialogue, spatial reasoning~\cite{quach2022non,shikra, gpt4roi,ferret, minigpt4v2, qwen_vl,macaw_llm,lisa}, and comprehensive video understanding~\cite{maaz2023video,videochat,vid2seq,chappa2024hatt,moviechat,embodiedgpt,videollava}.

Our research builds upon these advancements, proposing an innovative approach that seamlessly melds video analysis with language comprehension. By harnessing the strengths of large language models~\cite{liu2024visual}, our framework aims to bridge the gap between visual and textual information processing. This integration enables a more holistic interpretation of video content, paving the way for more accurate and context-aware applications.

A key challenge in this domain is the effective handling of streaming video inputs. Our work addresses this gap by developing comprehensive solutions that tackle issues of temporal alignment, long-context understanding, and real-time processing. We explore novel techniques in model architecture, training methodologies, and inference strategies to meet the unique demands of continuous video analysis.

Through this research, we not only contribute to the advancement of multimodal content analysis but also lay the groundwork for addressing complex tasks such as the analysis of tobacco-related content on social media platforms. Our approach demonstrates the potential of combining traditional video analysis techniques with cutting-edge language models to achieve a more nuanced and effective understanding of multimodal content across various domains.

\noindent \textbf{Tobacco Research:} Recently~\cite{kong2023understanding, murthy2023influence} used machine learning techniques to understand, specifically on YouTube, how the user profile attributes influence e-cigarette searches and analysis of their content and promotion strategies. Prior studies~\cite{ntad224} on e-cigarette content in social media have relied on qualitative and text-based machine learning methods. They developed an image-based computer vision model using a unique dataset of 6,999 Instagram images labeled for eight object classes related to e-cigarettes, demonstrating an innovative approach to automatically and scalably monitor tobacco promotions online. Murthy \emph{et al.}~\cite{ntad184} have relied on text-based analyses to detect e-cigarette content on social media, which addressed the visual nature of platforms like TikTok by developing a computer vision model that uses YOLOv7~\cite{wang2023yolov7}. This model was trained and tested on a dataset of 826 images, annotated for vaping devices, hands, and vapor clouds, extracted from 254 TikTok posts identified via relevant hashtags, showcasing a practical approach for monitoring e-cigarette content on visual social media platforms. Chappa et al.~\cite{chappa2024advanced} proposed a two-stage approach to process multi-modalities and analyze the tobacco promotion in the given videos.

\begin{figure*}[h!]
    \centering
    \includegraphics[width=\textwidth]{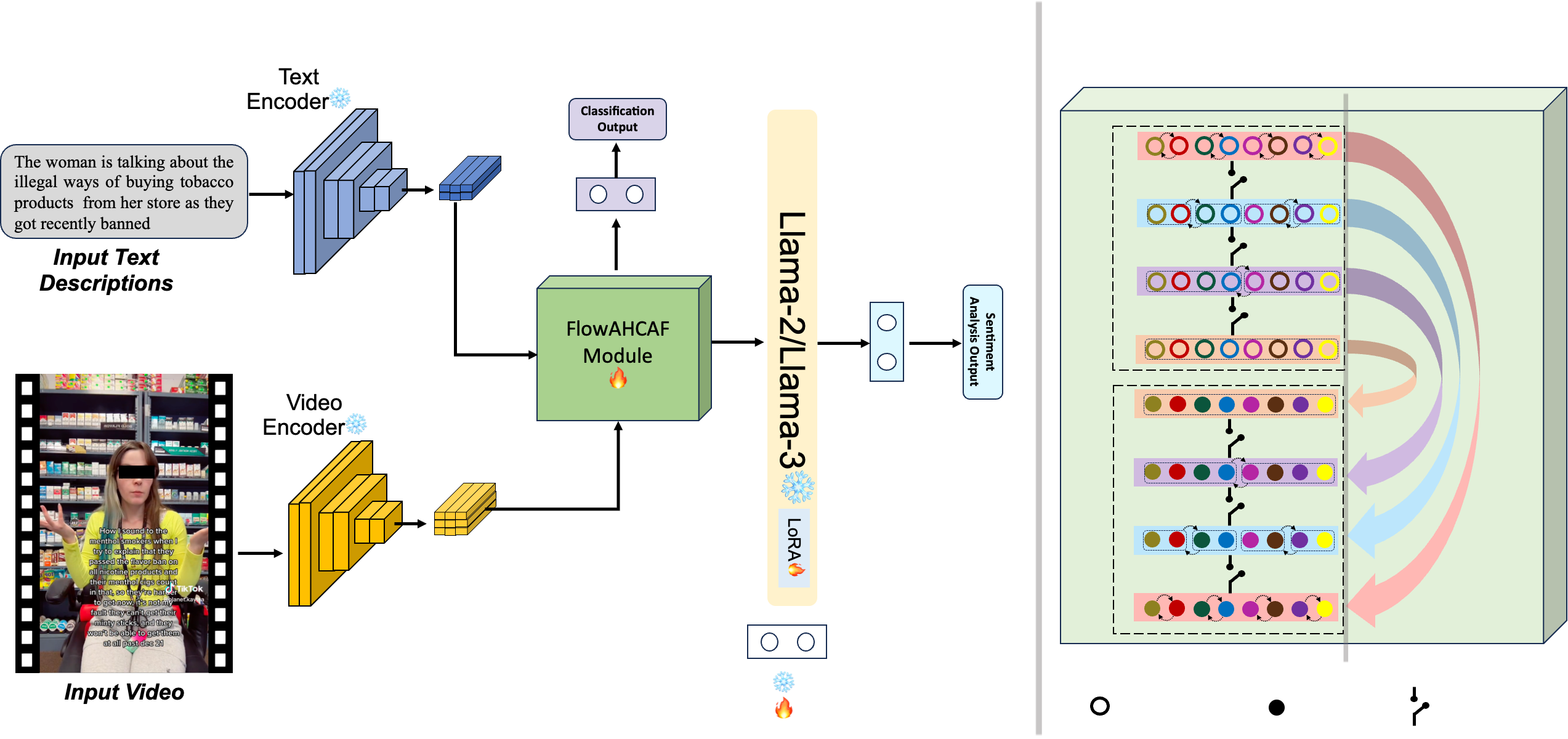}
    \put(-400,11){\small($\mathbf{V}$)}
    \put(-405,135){\small($\mathbf{T}$)}
    \put(-320,173){\small $\mathbf{T}_l$}
    \put(-320,50){\small $\mathbf{V}_l$}
    \put(-245,100){\small $\mathbf{O}$}
    \put(-214,30){\tiny \textbf{Task Specific}}
    \put(-214,24){\tiny \textbf{FC Layers}}
    \put(-223,15){\tiny \textbf{Frozen}}
    \put(-223,6){\tiny \textbf{Trainable}}
    \put(-143,171){\small \textbf{${l}_1$}}
    \put(-143,151){\small \textbf{${l}_2$}}
    \put(-143,131){\small \textbf{${l}_3$}}
    \put(-143,111){\small \textbf{${l}_4$}}
    \put(-143,94){\small \textbf{${l}_4$}}
    \put(-143,74){\small \textbf{${l}_3$}}
    \put(-143,54){\small \textbf{${l}_2$}}
    \put(-143,34){\small \textbf{${l}_1$}}
    \put(-112,164){\tiny \textbf{${\mathbf{G}_l}_1$}}
    \put(-92,159){\tiny \textbf{${\mathbf{G}_l}_2$}}
    \put(-112,144){\tiny \textbf{${\mathbf{G}_l}_2$}}
    \put(-92,139){\tiny \textbf{${\mathbf{G}_l}_3$}}
    \put(-112,124){\tiny \textbf{${\mathbf{G}_l}_3$}}
    \put(-92,119){\tiny \textbf{${\mathbf{G}_l}_4$}}
    \put(-112,89){\tiny \textbf{${\mathbf{G}_l}_4$}}
    \put(-92,84){\tiny \textbf{${\mathbf{G}_l}_3$}}
    \put(-112,69){\tiny \textbf{${\mathbf{G}_l}_3$}}
    \put(-92,64){\tiny \textbf{${\mathbf{G}_l}_2$}}
    \put(-112,49){\tiny \textbf{${\mathbf{G}_l}_2$}}
    \put(-92,44){\tiny \textbf{${\mathbf{G}_l}_1$}}
    \put(-17,171){\rotatebox{270}{\small \textbf{Flow Attention Mechanism}}}
    \put(-123,193){\small \textbf{FlowAHCAF Module}}
    \put(-131,12){\tiny\textbf{Textual}}
    \put(-131,6){\tiny\textbf{Features}}
    \put(-81,11){\tiny\textbf{Video}}
    \put(-81,5){\tiny\textbf{Features}}
    \put(-37,11){\tiny\textbf{Gating}}
    \put(-37,5){\tiny\textbf{Mechanism}}
    
    \caption{\textbf{Overall architecture of the  Flow-Attention Adaptive Semantic Hierarchical Fusion (FLAASH) framework}. The model takes text descriptions and video inputs, processes them through separate encoders, and fuses the information using the novel FlowAHCAF Module. The fused representation is then fed into a Llama-2/Llama-3 model with LoRA fine-tuning for various downstream tasks, including classification and sentiment analysis. Here, in the FlowAHCAF Module, we consider four hierarchy levels for our framework. The ~\cref{sec:Methodology} justifies all the math symbols. \textbf{Best viewed in color and zoom}.}
    \label{fig:flaash_framework}
\end{figure*}
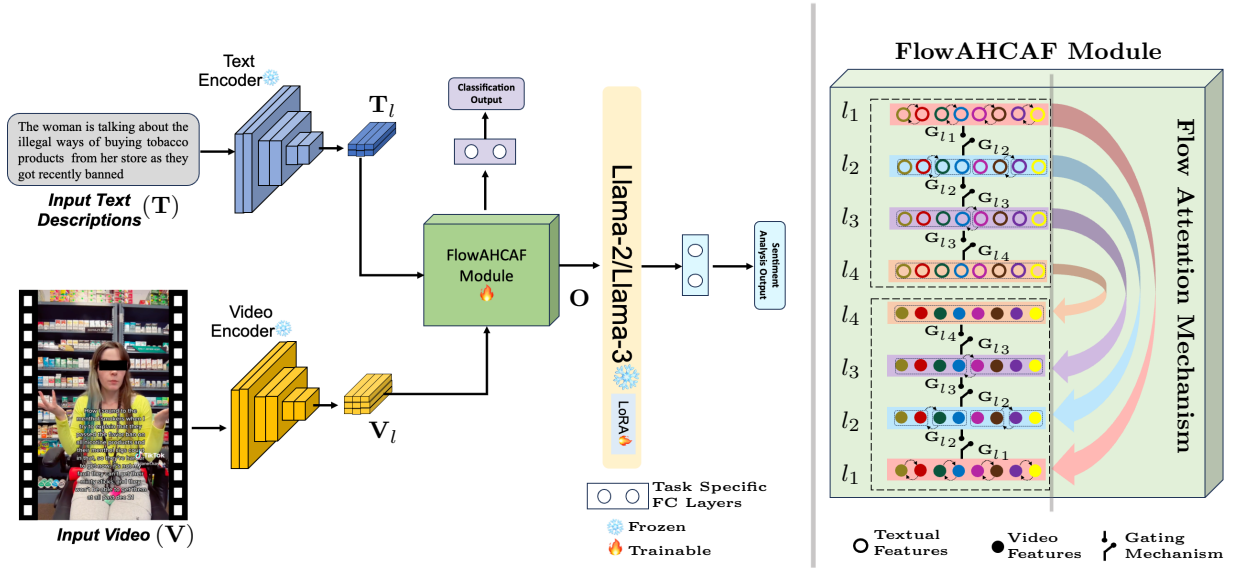

\section{Methodology}\label{sec:Methodology}

The Flow-Attention Adaptive Semantic Hierarchical Fusion (FLAASH) framework, as illustrated in \cref{fig:flaash_framework}, is designed to address the unique challenges of multimodal tobacco-related content analysis. FLAASH integrates flow network principles with hierarchical feature extraction and adaptive fusion mechanisms to overcome issues such as modality imbalance, temporal dynamics, contextual nuances, and scale variability in tobacco-related videos and associated text.

% \begin{figure}[t]
%     \centering
%     \includegraphics[width=\textwidth]{flaash_framework.png}
%     \caption{Overall architecture of the FLAASH framework. The model processes both text descriptions and video inputs through separate encoders, fuses the information using the novel FlowAHCAF Module, and utilizes a Llama-2/Llama-3 model with LoRA fine-tuning for various downstream tasks including classification and sentiment analysis. Snowflake symbols indicate frozen components, while the fire symbol represents trainable parts.}
%     \label{fig:flaash_framework}
% \end{figure}

\subsection{Input Processing}
The input processing stage of FLAASH is crucial for extracting meaningful features from both textual and visual modalities. This section details our approach to encoding text and video inputs, leveraging pre-trained models to capture rich, contextual representations. We describe the architecture and functionality of both the text and video encoders, highlighting their role in preparing data for the subsequent FlowAHCAF module.

\subsubsection{Text Encoder}
The text encoder in FLAASH is based on the BERT (Bidirectional Encoder Representations from Transformers)~\cite{devlin2018bert} architecture, a state-of-the-art model for natural language processing tasks. Our implementation consists of $L$ transformer layers, where each layer $l$ produces features $\mathbf{T}_l \in \mathbb{R}^{S \times D_l}$. Here, $S$ represents the sequence length of the input text, and $D_l$ is the feature dimension at layer $l$.

We leverage a pre-trained BERT model to capitalize on its robust language understanding capabilities. The encoder processes input text sequences, capturing contextual information and semantic relationships crucial for understanding tobacco-related content. Each transformer layer progressively refines the text representation, with lower layers capturing more local syntactic information and higher layers encoding more global semantic concepts.

To maintain computational efficiency and prevent overfitting on our specific task, we keep the text encoder frozen during training. This approach allows us to benefit from the rich language representations learned from large-scale pre-training while adapting the subsequent layers of FLAASH to our specific tobacco content analysis task.

\subsubsection{Video Encoder}
For processing video inputs, we employ a 3D ResNet-based architecture, which has shown exceptional performance in action recognition and video understanding tasks. Our video encoder comprises $L$ levels, where each level $l$ generates features $\mathbf{V}_l \in \mathbb{R}^{T_l \times C_l}$. In this representation, $T_l$ denotes the temporal dimension, capturing the dynamic nature of video content, and $C_l$ represents the number of channels at level $l$.

The 3D ResNet architecture is particularly suited for our task as it effectively captures both spatial and temporal information in video frames. Lower levels of the encoder focus on local spatiotemporal patterns, such as texture and short-term motion, while higher levels abstract these into more complex representations of objects, actions, and scenes relevant to tobacco use and promotion.

Similar to the text encoder, we utilize a pre-trained 3D ResNet model and keep it frozen during training. This strategy allows us to leverage robust visual representations learned from large-scale video datasets while avoiding the need for extensive fine-tuning on our relatively smaller tobacco-related dataset.

By employing these sophisticated, pre-trained encoders for both text and video inputs, FLAASH establishes a strong foundation for multi-modal fusion. The rich, contextual representations extracted by these encoders provide the subsequent FlowAHCAF module with high-quality features, enabling more effective integration of visual and textual information for nuanced tobacco content analysis.

\subsection{FlowAHCAF Module}
The Flow-Attention Adaptive Hierarchical Cross-Attention Fusion (FlowAHCAF) module is the core component of our framework. It is designed to effectively fuse multimodal information across multiple levels of abstraction. This module integrates flow-attention mechanisms, hierarchical processing, gating, and adaptive weighting to capture complex relationships between video and text inputs.

\subsubsection{Hierarchical Structure}
The FlowAHCAF module is built on a hierarchical structure with $L$ levels (default $L=3$), allowing it to process information at multiple scales:

% \begin{itemize}
Each level $l$ operates on increasingly abstract representations of the input:
    \begin{equation}
        \mathbf{V}_l = f_V^l(\mathbf{V}_{l-1}), \quad \mathbf{T}_l = f_T^l(\mathbf{T}_{l-1})
    \end{equation}
where $f_V^l$ and $f_T^l$ are level-specific transformation functions for video and text features, respectively. Lower levels capture fine-grained details, while higher levels focus on more abstract, global information.

This hierarchical approach forms the backbone of our module, upon which we apply our flow-attention mechanism, gating, and adaptive weighting.

\subsubsection{Flow-Attention Mechanism}
At each hierarchical level, we apply a flow-attention mechanism inspired by flow network theory. This mechanism implements flow conservation on both source and sink aspects, avoiding trivial attention patterns:
\begin{equation}
    \mathbf{F}_l = \text{FlowAttention}(\mathbf{V}_l, \mathbf{T}_l)
\end{equation}
The flow computation for each level $l$ is given by:
\begin{equation}
    I_i^l = \phi(\mathbf{Q}_i^l) \sum_{j=1}^{m} \phi(\mathbf{K}_j^l)^{\sf T}, \quad O_j^l = \phi(\mathbf{K}_j^l) \sum_{i=1}^{n} \phi(\mathbf{Q}_i^l)^{\sf T}
\end{equation}
\begin{equation}
    \text{FlowAttention} = \frac{\phi(\mathbf{Q}^l)}{\mathbf{I}^l} \left( \phi(\mathbf{K}^l)^{\sf T} (\text{Softmax}(\widehat{\mathbf{O}}^l) \odot \mathbf{V}^l) \right)
\end{equation}
where $\phi(\cdot) = \text{softplus}(\cdot) + \epsilon$ is a non-negative element-wise non-linear projection, with $\epsilon = 10^{-6}$ for numerical stability. The query, key, and value projections for each level are learned as:
\begin{equation}
    \mathbf{Q}^l = W^Q_l \mathbf{T}_l, \quad \mathbf{K}^l = W^K_l \mathbf{V}_l, \quad \mathbf{V}^l = W^V_l \mathbf{V}_l
\end{equation}
\subsubsection{Gating Mechanism}
We implement a gating mechanism to control the information flow between modalities at each hierarchical level. For each level $l$, the gating function $G_l$ is defined as:
\begin{equation}
    \mathbf{G}_l = \sigma(W_g^l[\mathbf{F}_l; \mathbf{T}_l] + b_g^l)
\end{equation}
where $\sigma$ is the sigmoid activation function, $W_g^l$ and $b_g^l$ are level-specific learnable parameters, and $[\cdot;\cdot]$ denotes concatenation. The gated features for each level are then computed as:
\begin{equation}
    \mathbf{F}_l^{gated} = \mathbf{G}_l \odot \mathbf{F}_l
\end{equation}
This gating mechanism allows the model to emphasize relevant features at each level of the hierarchy selectively.

\subsubsection{Adaptive Weighting}
We introduce an adaptive weighting mechanism to adjust the importance of different hierarchical levels dynamically. For each level $l$, we have:
\begin{equation}
    \mathbf{H}_l = \alpha_l \mathbf{F}_l^{gated} + (1 - \alpha_l) \mathbf{V}_l
\end{equation}
where $\alpha_l$ are learnable parameters initialized to 0.5. It allows the model to find an optimal balance between the fused information and the original video features at each hierarchical level.

The final fused representation is obtained by combining the adaptively weighted features from all levels:
\begin{equation}
    \mathbf{O} = \text{Fusion}([\mathbf{H}_1; \mathbf{H}_2; ...; \mathbf{H}_L])
\end{equation}
This fusion operation learns to combine information across all hierarchical levels, producing a rich, multi-scale representation of the input video and text.

Through this hierarchical structure, integrated with flow attention, gating, and adaptive weighting, FlowAHCAF effectively processes and fuses multimodal information at multiple scales. This approach enables the capture of both local details and global context, which is crucial for nuanced tobacco-related content analysis.

\subsection{Llama-2/Llama-3 Integration}
The fused representation from the FlowAHCAF module is fed into a Llama-2-7B-Chat~\cite{touvron2023llama} or Llama-3-8B-Instruct~\cite{meta2024introducing} large language model for further processing. This integration allows our framework to leverage the powerful language understanding capabilities of these models for tobacco-related content analysis. These advanced language models bring several advantages to our framework. They excel at understanding complex contexts, which is crucial for interpreting nuanced tobacco-related content. Their ability to adapt to new tasks with minimal fine-tuning enhances the flexibility of our system. Moreover, these models can handle various language tasks simultaneously, supporting our multi-faceted analysis approach. The choice between Llama-2 and Llama-3 depends on the specific requirements of the analysis task and computational resources available, with Llama-3 generally offering improved performance but potentially requiring more computational power.

\subsubsection{LoRA Fine-tuning}
We employ Low-Rank Adaptation (LoRA)~\cite{hu2021lora} to fine-tune the Llama model efficiently. This technique allows us to adapt the pre-trained model to our specific task domain while maintaining its general language understanding capabilities. LoRA works by inserting trainable rank decomposition matrices into each layer of the Llama model, offering significant benefits in efficiency, performance, and flexibility. It reduces the number of trainable parameters, making fine-tuning faster and less resource-intensive, while often achieving comparable or better performance than full fine-tuning. This approach allows for task-specific adaptations without altering the base model, facilitating multi-task learning. Our implementation of LoRA involves identifying key layers in the Llama model for adaptation, initializing low-rank matrices for these layers, and training these matrices on our tobacco-related dataset while keeping the base model frozen. This approach enables us to tailor the powerful Llama model to our specific tobacco content analysis tasks without losing its broader language understanding capabilities.

\subsection{Task-Specific Outputs}
The output from the Llama model is then passed through task-specific fully connected (FC) layers to produce outputs for classification and sentiment analysis tasks related to tobacco content. This final stage of our pipeline is crucial for translating the rich, contextual representations from Llama into actionable insights. Our task-specific layers include a multi-class FC layer for categorizing content into predefined tobacco-related categories, an FC layer with multiple outputs for sentiment analysis corresponding to different sentiment dimensions relevant to tobacco use, and an LSTM-based layer that ensures consistency in predictions across video frames. These layers are trained end-to-end with the rest of the model, allowing them to learn task-specific features while benefiting from the rich representations provided by the Llama model and FlowAHCAF module. To enhance the model's performance on these specific tasks, we employ Focal Loss for classification to address potential class imbalance in tobacco-related categories, a custom sentiment loss that captures the nuanced nature of attitudes towards tobacco use, and a temporal consistency loss to ensure coherent predictions across video frames. By combining the powerful language understanding of Llama models with these carefully designed task-specific layers, FLAASH can provide detailed, nuanced analysis of tobacco-related content across various dimensions, supporting more informed public health strategies and interventions.

\subsection{Multi-Task Learning Objective}

To effectively capture the diverse aspects of tobacco-related content analysis, we employ a multi-task learning approach. This approach enables our model to simultaneously learn and perform various tasks relevant to understanding tobacco usage, sentiment, and context in multimodal data. Our composite loss function is designed to address classification, sentiment analysis, temporal consistency, feature discrimination, and regularization, defined as follows:
\begin{equation}
    \mathcal{L}_{\text{total}} = \lambda_1 \mathcal{L}_{\text{cls}} + \lambda_2 \mathcal{L}_{\text{sent}} + \lambda_3 \mathcal{L}_{\text{temp}} + \lambda_4 \mathcal{L}_{\text{cont}} + \lambda_5 \mathcal{L}_{\text{reg}}
\end{equation}
where $\lambda_1, \lambda_2, \lambda_3, \lambda_4, \lambda_5$ are weighting coefficients for each loss component.

The \textbf{classification loss ($\mathcal{L}_{\text{cls}}$)} focuses on categorizing tobacco products and usage scenarios, utilizing a cross-entropy formulation:
\begin{equation}
    \mathcal{L}_{\text{cls}} = -\sum_{i=1}^{N} \sum_{c=1}^{C} y_{i,c} \log(\hat{y}_{i,c})
\end{equation}
Here, $N$ represents the number of samples, $C$ the number of classes, $y_{i,c}$ the true label, and $\hat{y}_{i,c}$ the predicted probability.

For analyzing sentiment towards tobacco use, we employ a \textbf{sentiment analysis loss ($\mathcal{L}_{\text{sent}}$)}, structured similarly to the classification loss:
\begin{equation}
    \mathcal{L}_{\text{sent}} = -\sum_{i=1}^{N} \sum_{s=1}^{S} y_{i,s} \log(\hat{y}_{i,s})
\end{equation}
where $S$ denotes the number of sentiment categories.

To ensure consistency in predictions across video frames, we incorporate a \textbf{temporal consistency loss ($\mathcal{L}_{\text{temp}}$)}:
\begin{equation}
    \mathcal{L}_{\text{temp}} = \frac{1}{T-1} \sum_{t=1}^{T-1} \|\mathbf{f}_t - \mathbf{f}_{t+1}\|_2^2
\end{equation}
In this formulation, $\mathbf{f}_t$ represents the features at time step $t$, and $T$ is the total number of time steps.

To enhance the discrimination between different tobacco-related concepts in the embedding space, we employ a \textbf{contrastive loss ($\mathcal{L}_{\text{cont}}$)}:
\begin{equation}
    \mathcal{L}_{\text{cont}} = -\log \frac{\exp(s_p / \tau)}{\exp(s_p / \tau) + \sum_{n=1}^{N} \exp(s_n / \tau)}
\end{equation}
Here, $s_p$ is the similarity score for positive pairs, $s_n$ for negative pairs, $\tau$ is a temperature parameter, and $N$ is the number of negative samples.

Finally, to prevent overfitting and encourage sparsity in the attention weights, we include a \textbf{regularization loss ($\mathcal{L}_{\text{reg}}$)}:
\begin{equation}
    \mathcal{L}_{\text{reg}} = \|\Theta\|_2^2 + \|\mathbf{A}\|_1
\end{equation}
where $\Theta$ represents the model parameters and $\mathbf{A}$ the attention weights.

This comprehensive multi-task loss function is designed to classify tobacco products and usage scenarios accurately, capture the sentiment associated with tobacco-related content, ensure temporal coherence in video analysis, learn discriminative features for different tobacco-related concepts, and prevent overfitting while encouraging meaningful attention patterns. Simultaneously optimizing for these diverse objectives, our model holistically understands tobacco-related content across various modalities and temporal contexts.

% The weighting coefficients $\lambda_1, \lambda_2, \lambda_3, \lambda_4, \lambda_5$ are hyperparameters that control the contribution of each loss component. These are tuned on a validation set to optimize the overall performance of the model on tobacco-related content analysis tasks.

% \subsubsection{Implementation Details}
% We implement FlowAHCAF using PyTorch 1.9.0. The video encoder is based on a 3D ResNet-101 architecture, pre-trained on Kinetics-400, and fine-tuned on our dataset. The text encoder utilizes a BERT-base-uncased model, also fine-tuned on our dataset. We use Adam optimizer with a learning rate of 1e-4, weight decay of 1e-5, and train the model for 100 epochs on 4 NVIDIA V100 GPUs with 32GB memory each. We employ a linear warm-up strategy for the first 10 epochs and a cosine annealing schedule thereafter. The batch size is set to 32, with each sample consisting of 32 video frames and up to 512 text tokens.

% \textcolor{red}{to elaborate, include comparison of different fusion techniques}

\section{Experiments}

This section details our comprehensive evaluation of the FLAASH model. We begin by describing our experimental setup, including the datasets used, implementation details, and evaluation metrics. We then present state-of-the-art comparisons on both tobacco-specific content analysis and general video question answering tasks. Finally, we conduct a series of ablation studies to analyze the impact of various components of our model and provide qualitative insights into its performance.

\subsection{Experimental Setup}

\subsubsection{Datasets}
We evaluate the proposed FLAASH model on multiple datasets to assess its performance and generalization capabilities. Our primary dataset is the Multimodal Tobacco Content Analysis Dataset (MTCAD), which comprises 5,730 videos from YouTube and TikTok, focusing on tobacco-related content. This dataset includes diverse video types such as product reviews, personal vlogs, promotional material, and public health messages, annotated with metadata including user engagement metrics, tobacco product types, and sentiment labels.
To evaluate the generalization capability of our model and its effectiveness in multi-modal understanding beyond tobacco-specific content, we use several Video Question Answering datasets. These include MSVD-QA~\cite{xu2017video} with 50,505 question-answer pairs for 1,970 short video clips, MSR-VTT-QA~\cite{xu2017video} containing 243,680 question-answer pairs for 10,000 video clips across various categories, and ActivityNet-QA~\cite{caba2015activitynet} which includes 58,000 question-answer pairs for 5,800 videos focusing on human activities and events.
For MTCAD, we use a 70:15:15 split for training, validation, and test sets. For the VQA datasets, we follow the standard splits provided by their respective authors.
\subsubsection{Implementation Details}
We implement FLAASH using PyTorch 1.9.0. Our model architecture consists of a Video Encoder based on a 3D ResNet-101 architecture, pre-trained on Kinetics-400 and fine-tuned on our datasets, and a Text Encoder using a BERT-base-uncased model, also fine-tuned on our datasets. Training is performed using Adam optimizer with a learning rate of 1e-4 and weight decay of 1e-5. We train for 100 epochs on 4 NVIDIA V100 GPUs, using a batch size of 32, with each sample consisting of 32 video frames and up to 512 text tokens.
\subsubsection{Evaluation Metrics}
We use a range of metrics to evaluate our model's performance. These include Classification Accuracy for tobacco product and usage scenario classification on MTCAD, Sentiment Analysis F1-Score to evaluate sentiment understanding on MTCAD, Temporal Consistency Score (TCS) to assess the model's ability to maintain consistent predictions across video frames, Multi-modal Fusion Quality (MFQ) to quantify the effectiveness of our FlowAHCAF fusion mechanism, and Answer Accuracy for evaluating performance on VQA datasets. These metrics are further explained in~\cref{appendix:detailed metrics}.
\subsection{State-of-the-Art Comparisons}
This subsection presents our model's performance in comparison to existing state-of-the-art methods. We evaluate FLAASH on both tobacco-specific content analysis and general video question answering tasks to demonstrate its effectiveness and versatility.
\subsubsection{Tobacco Content Analysis Performance}
The performance comparison on the MTCAD dataset in~\cref{tab:mtcad_results} demonstrates FLAASH's superior effectiveness in analyzing tobacco-related content across critical metrics. FLAASH achieves the highest classification accuracy of 0.83, surpassing VideoFormer (0.80) by 3\%, indicating an enhanced ability to categorize tobacco-related content accurately. In sentiment analysis, FLAASH's F1 score of 0.79 shows a significant 4\% improvement over VideoFormer (0.75), suggesting better comprehension of emotional context.

FLAASH's temporal consistency score of 0.75 outperforms other methods, with VideoFormer following at 0.72. This highlights FLAASH's capacity to maintain consistent predictions across video frames. FLAASH's multi-modal fusion quality score of 0.81, compared to VideoFormer's 0.77, underscores its effectiveness in integrating visual and textual information.

The performance progression from simpler models like YOLO + BERT to more complex ones like 3D-CNN + LSTM and VideoFormer shows a clear improvement trend. However, FLAASH's consistent outperformance across all metrics emphasizes its advanced capabilities in handling multi-modal tobacco-related content analysis. These results validate the effectiveness of FLAASH's novel architecture, particularly its Flow-Attention Adaptive Hierarchical Cross-Attention Fusion mechanism and multi-task learning approach, in addressing the unique challenges of analyzing tobacco-related content on social media platforms.
\vspace{-4mm}
\begin{table}[ht]
\centering
\caption{Performance comparison on MTCAD dataset. Cls. Acc.: Classification Accuracy; Sent. F1: Sentiment Analysis F-1 score; TCS: Temporal Consistency Score; MFQ: Multi-modal Fusion Quality.}
\label{tab:mtcad_results}
\begin{tabularx}{\columnwidth}{Lcccc}
\hline
Method & Cls. Acc. & Sent. F1 & TCS & MFQ \\
\hline
YOLO + BERT~\cite{devlin2018bert} & 0.75 & 0.70 & 0.68 & - \\
3D-CNN + LSTM~\cite{hochreiter1997long} & 0.78 & 0.73 & 0.71 & - \\
VideoFormer~\cite{ge2022bridging} & 0.80 & 0.75 & 0.72 & 0.77 \\
\textbf{FLAASH (Ours)} & \textbf{0.83} & \textbf{0.79} & \textbf{0.75} & \textbf{0.81} \\
\hline
\end{tabularx}
\end{table}

%\vspace{-4mm}
\subsubsection{Generalization to Video Question Answering}
The results presented in~\cref{tab:vqa_results} demonstrate FLAASH's competitive performance across three standard video question-answering datasets: MSVD-QA, MSRVTT-QA, and ActivityNet-QA (AN-QA). These datasets serve as benchmarks for evaluating the generalization capabilities of multi-modal models beyond tobacco-specific content analysis.

On the MSVD-QA dataset, FLAASH achieves state-of-the-art performance with an answer accuracy of 73.5\% and a score of 4.0, surpassing the previous best model, VideoGPT+~\cite{maaz2024videogpt+}, by 1.1 percentage points. This improvement suggests FLAASH's enhanced ability to understand and reason about general video content and associated questions.

For the MSRVTT-QA dataset, FLAASH performs competitively, achieving an accuracy of 58.7\% and a score of 3.4. While slightly below VideoGPT+'s performance (60.6\% / 3.6), FLAASH maintains a strong position among top-performing models, outperforming many recent approaches such as Video-LLaVA~\cite{lin2023video} and VideoChat2~\cite{li2024mvbench}.

On the challenging ActivityNet-QA dataset, FLAASH sets a new state-of-the-art with an accuracy of 51.2\% and a score of 3.7, marginally outperforming VideoGPT+ (50.6\% / 3.6). This result is particularly noteworthy given ActivityNet-QA's focus on complex human activities and events, demonstrating FLAASH's robustness in understanding diverse and dynamic video content.

The consistent high performance across these datasets highlights FLAASH's versatility and generalization capabilities. Starting from FrozenBiLM~\cite{yang2022zero} to recent models like VideoGPT+, there is a clear trend of improvement in video understanding and question-answering tasks. FLAASH builds upon this progress, leveraging its Flow-Attention Adaptive Hierarchical Cross-Attention Fusion mechanism to achieve superior or competitive results across diverse video understanding tasks.

These results validate FLAASH's effectiveness in its primary domain of tobacco-related content analysis and underscore its potential as a general-purpose video understanding model. The strong performance on standard VQA benchmarks suggests that the innovations introduced in FLAASH, particularly its multi-modal fusion techniques and hierarchical processing, offer benefits that extend beyond specialized domains to general video comprehension tasks.

\begin{table*}[t]
\centering
\caption{Results on VQA datasets (Answer Accuracy / Score)}
\label{tab:vqa_results}
% \small
\begin{tabular}{lccc}
\hline
Model & MSVD-QA & MSRVTT-QA & AN-QA \\
\hline
FrozenBiLM~\cite{yang2022zero} & 32.2 / - & 16.8 / - & 24.7 / - \\
VideoChat~\cite{li2023videochat} & 56.3 / 2.8 & 45.0 / 2.5 & 26.5 / 2.2 \\
LLaMA Adapter~\cite{zhang2023llama} & 54.9 / 3.1 & 43.8 / 2.7 & 34.2 / 2.7 \\
Video-LLaMA~\cite{zhang2023video} & 51.6 / 2.5 & 29.6 / 1.8 & 12.4 / 1.1 \\
Video-ChatGPT~\cite{maaz2023video} & 64.9 / 3.3 & 49.3 / 2.8 & 35.2 / 2.8 \\
ChatUniVi~\cite{jin2024chat} & 65.0 / 3.6 & 54.6 / 3.1 & 45.8 / 3.2 \\
LLaMA-VID~\cite{li2023llama} & 70.0 / 3.7 & 58.9 / 3.3 & 47.5 / 3.3 \\
Video-LLaVA~\cite{lin2023video} & 70.7 / 3.9 & 59.2 / 3.5 & 45.3 / 3.3 \\
VideoChat2~\cite{li2024mvbench} & 70.0 / 3.9 & 54.1 / 3.3 & 49.1 / 3.3 \\
VideoGPT+~\cite{maaz2024videogpt+} & 72.4 / 3.9 & \textbf{60.6 / 3.6} & 50.6 / 3.6 \\
\textbf{FLAASH (Ours)} & \textbf{73.5 / 4.0} & 58.7 / 3.4 & \textbf{51.2 / 3.7} \\
\hline
\end{tabular}
\end{table*}

\subsection{Ablation Studies}

In this subsection, we present a series of ablation studies to analyze the impact of various components of our FLAASH model. We examine the effectiveness of the FlowAHCAF mechanism, investigate the contribution of its individual components, and assess the benefits of our multi-task learning approach.

\subsubsection{Impact of FlowAHCAF}

The ablation study in~\cref{tab:fusion_comparison} demonstrates FlowAHCAF's superiority over baseline fusion techniques on the MTCAD dataset. FlowAHCAF consistently outperforms both simple concatenation and cross-modal attention across all metrics. Compared to concatenation, FlowAHCAF shows substantial improvements: 11\% in accuracy (0.83 v.s 0.72), 11\% in F1 score (0.79 v.s 0.68), 10\% in temporal consistency (0.75 v.s 0.65), and 11\% in multi-modal fusion quality (0.81 v.s 0.70). Against cross-modal attention, FlowAHCAF still achieves notable gains: 5\% in accuracy, 5\% in F1 score, 4\% in temporal consistency, and 5\% in fusion quality. These results validate FlowAHCAF's effectiveness in integrating visual and textual information, capturing complex inter-modal relationships, and maintaining temporal coherence. The consistent improvements across all metrics underscore the efficacy of FlowAHCAF's design, particularly its use of flow network principles and hierarchical processing, in addressing the challenges of multi-modal fusion for tobacco-related content analysis.

\begin{table}[ht]
\centering
\caption{Comparison of fusion techniques on MTCAD}
\label{tab:fusion_comparison}
\begin{tabularx}{\columnwidth}{Lcccc}
\hline
Method & Acc. & F1 & TCS & MFQ \\
\hline
Concatenation~\cite{maaz2023video} & 0.72 & 0.68 & 0.65 & 0.70 \\
Cross-Modal Attention~\cite{tan2024koala} & 0.78 & 0.74 & 0.71 & 0.76 \\
\textbf{FlowAHCAF (Ours)} & \textbf{0.83} & \textbf{0.79} & \textbf{0.75} & \textbf{0.81} \\
\hline
\end{tabularx}
\end{table}
%\vspace{-4mm}

\subsubsection{Analysis of FlowAHCAF Components}

The ablation study in~\cref{tab:flowahcaf_components} reveals the incremental benefits of each component in the FlowAHCAF mechanism. Starting from a base model without FlowAHCAF, we observe consistent improvements as each component is added. Flow attention provides a significant boost, increasing accuracy from 0.76 to 0.79 and improving other metrics similarly. The addition of gating further enhances performance, with accuracy rising to 0.81. Adaptive weighting contributes additional gains, bringing accuracy to 0.82. The full FlowAHCAF, combining all components, achieves the best performance across all metrics (Acc: 0.83, F1: 0.79, TCS: 0.75, MFQ: 0.81). This stepwise improvement demonstrates the synergistic effect of these mechanisms in multi-modal fusion. Each component addresses specific aspects of the fusion process: flow attention captures complex interactions, gating selectively emphasizes relevant features, and adaptive weighting balances the importance of different hierarchical levels. The cumulative 7\% improvement in accuracy from the base model to full FlowAHCAF underscores the effectiveness of this multi-component approach in enhancing multi-modal understanding for tobacco-related content analysis.
\begin{table}[ht]
\centering
\caption{Ablation study of FlowAHCAF components on MTCAD dataset}
\label{tab:flowahcaf_components}
\begin{tabularx}{\columnwidth}{Xcccc}
\hline
Configuration & Acc. & F1 & TCS & MFQ \\
\hline
Base model (w/o FlowAHCAF) & 0.76 & 0.72 & 0.69 & 0.74 \\
+ Flow Attention & 0.79 & 0.75 & 0.71 & 0.77 \\
+ Gating & 0.81 & 0.77 & 0.73 & 0.79 \\
+ Adaptive Weighting & 0.82 & 0.78 & 0.74 & 0.80 \\
\textbf{Full FlowAHCAF} & \textbf{0.83} & \textbf{0.79} & \textbf{0.75} & \textbf{0.81} \\
\hline
\end{tabularx}
\end{table}
%\vspace{-4mm}

\subsubsection{Multi-Task Learning Analysis}

The ablation study in~\cref{tab:multitask_ablation} demonstrates the effectiveness of our multi-task learning approach across MTCAD and two standard VQA datasets. Starting with classification only, we observe steady improvements as additional components are incorporated. On MTCAD, adding sentiment analysis increases accuracy from 0.80 to 0.82, while temporal consistency further boosts it to 0.83. The complete model, including contrastive and regularization losses, achieves the highest accuracy of 0.85. Similar trends are observed on MSVD-QA and MSRVTT-QA, with the full model outperforming partial implementations across all datasets. The consistent improvements (5\% on MTCAD, 6\% on MSVD-QA, and 5\% on MSRVTT-QA from classification-only to full model) highlight the complementary nature of these components. Each task contributes to a more comprehensive understanding of the video content, with sentiment capturing emotional context, temporal consistency ensuring coherent predictions, and the complete model leveraging feature discrimination and regularization for optimal performance. This multi-task approach enhances performance on the tobacco-specific MTCAD dataset and demonstrates improved generalization capabilities on standard VQA tasks, underscoring its versatility in video understanding.
\begin{table}[ht]
\centering
\caption{Ablation of multi-task learning components across datasets}
\label{tab:multitask_ablation}
% \small
\begin{tabularx}{\columnwidth}{Lccc}
\hline
Components & MTCAD & MSVD-QA & MSRVTT-QA \\
\hline
Classification only & 0.80 & 0.68 & 0.56 \\
+ Sentiment & 0.82 & 0.70 & 0.58 \\
+ Temporal & 0.83 & 0.72 & 0.60 \\
\textbf{All of them (Ours)} & \textbf{0.85} & \textbf{0.74} & \textbf{0.61} \\
\hline
\end{tabularx}
\end{table}
\vspace{-4mm}

\subsection{Qualitative Analysis}

\begin{figure*}[ht]
    \centering
    \includegraphics[width=0.8\textwidth]{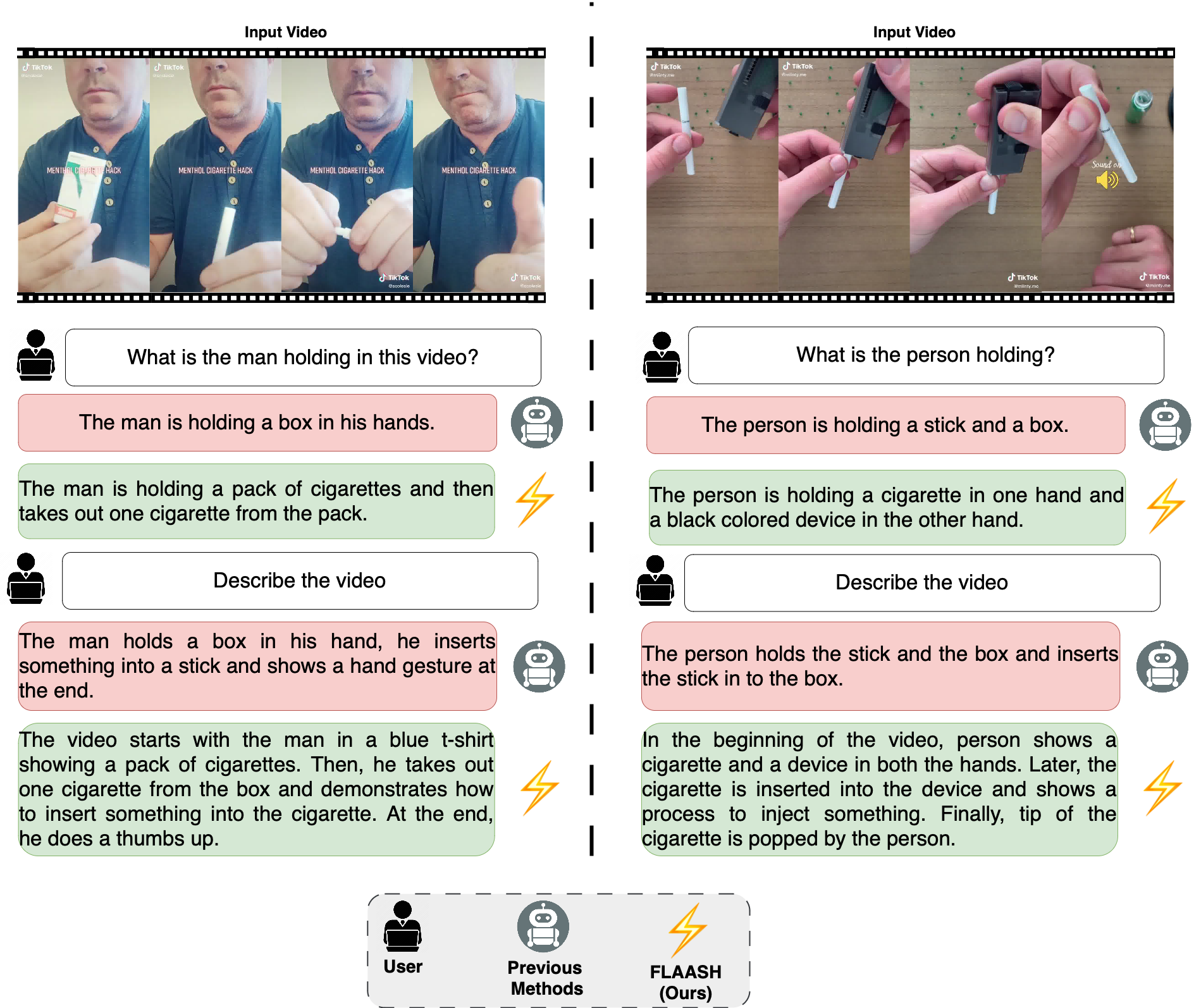}
    \caption{Comparison of video content analysis between previous methods~\cite{maaz2023video, tan2024koala} and FLAASH (Ours). We present the analysis for two different videos with similar prompts. \textbf{Best viewed in color and zoom.}}
    \label{fig:results}
\end{figure*}

\cref{fig:results} presents a comparative analysis of our FLAASH model against previous methods~\cite{maaz2023video, tan2024koala} for different types of tobacco-related content. The figure showcases two distinct scenarios: one depicting a man handling cigarettes and another showing hands manipulating a cigarette and an associated device. This qualitative analysis reveals FLAASH's ability to: (1) Focus on relevant visual elements while maintaining awareness of the broader context. (2) Integrate textual queries seamlessly with visual information for more precise responses. (3) Capture fine-grained details that are often missed by previous methods. (4) Provide more coherent and contextually relevant descriptions of tobacco-related activities. The stark contrast between FLAASH's responses and those of previous methods underscores the effectiveness of our FlowAHCAF mechanism in achieving a more holistic understanding of multi-modal tobacco-related content. This improved comprehension is crucial for accurate analysis and interpretation of tobacco use and promotion in social media videos, supporting more informed public health strategies and interventions.

\section{Discussion}
\subsection{Advantages in Tobacco-Related Content Analysis}

The Flow-Attention mechanism, as a critical component of FlowAHCAF, offers several advantages in the context of tobacco-related content analysis:

\textbf{Balanced Feature Interaction}: By treating attention as a flow problem, we ensure a more balanced interaction between video and text features. It is particularly important when analyzing tobacco products alongside related textual descriptions or comments in videos.
    
\textbf{Contextual Understanding}: The competition and allocation processes help capture the contextual relationships between visual tobacco product representations and associated textual information, enabling a more nuanced understanding of the content.
    
\textbf{Robustness to Noise}: The flow conservation principle helps filter out irrelevant or noisy information, which is crucial when dealing with diverse user-generated content related to tobacco use.
    
\textbf{Adaptive Focus}: The mechanism allows the model to adaptively focus on the most relevant aspects of video and text data, essential for analyzing various types of tobacco-related content, from product reviews to health awareness videos.

\subsection{Why FlowAttention is required?}

\begin{enumerate}
    \item \textbf{Enhanced Interaction Between Modalities}: The Flow-Attention mechanism allows for more sophisticated interaction between text and video data by leveraging the concepts of flow conservation and competition. It can help better capture the relationships and dependencies between the two modalities.

\item \textbf{Avoiding Trivial Attention}: Traditional attention mechanisms can sometimes produce trivial attention weights, especially in linear Transformers. Flow-Attention introduces a competition mechanism among tokens, which helps avoid trivial attention by ensuring that only the most relevant tokens are attended to.

\item \textbf{Global Context Understanding}: The flow network perspective helps understand the global context by considering each node's incoming and outgoing flow (token). It can provide valuable insights into how information flows through the network, leading to a more informed and contextually relevant fusion of text and video features.

\item \textbf{Scalability and Efficiency}: Flow-Attention achieves linear complexity in computing attention weights by leveraging flow conservation principles. It makes it scalable and efficient, especially when dealing with large datasets and long sequences, common in multi-modal learning scenarios.

\item \textbf{Non-Trivial Re-weighting}: The competition mechanism in Flow-Attention ensures that the attention weights are non-trivially re-weighted based on the flow capacity. It helps highlight the most critical features of text and video inputs, leading to more accurate and meaningful fusion.

\item \textbf{Generalization}: Flow-Attention does not rely on specific inductive biases, making it more general and adaptable to various data types and tasks. It can be beneficial in the context of MTCAD, where the nature of the data and the analysis tasks can vary significantly.
\end{enumerate}

\section{Conclusions}
This paper introduced FLAASH, a novel multi-modal framework for analyzing tobacco-related content in social media videos. Our FlowAHCAF mechanism demonstrated superior performance in fusing visual and textual information, outperforming baseline techniques across all metrics. The multi-task learning approach, incorporating classification, sentiment analysis, and temporal consistency, significantly improved tobacco-specific and standard VQA datasets.

FLAASH's state-of-the-art results highlight its potential as a valuable tool for researchers and policymakers to understand tobacco-related content on social media. The framework's versatility suggests promising applications in broader public health contexts and other domains requiring sophisticated multi-modal analysis.

In conclusion, FLAASH represents a significant advance in applying AI to public health challenges, particularly in addressing tobacco promotion in the digital age. As social media's influence grows, such tools will be crucial in developing practical, data-driven public health strategies.

% \vspace{-1mm}
\noindent\textbf{Limitations:}
The FLAASH framework, while innovative in its approach to tobacco-related content analysis, is not without limitations. Its computational complexity, stemming from the hierarchical FlowAHCAF module and the integration of large language models like Llama-2/Llama-3, may hinder real-time processing capabilities, especially for longer videos. The heavy reliance on pre-trained components, including text and video encoders, could introduce inherited biases or limitations. While using LoRA for fine-tuning is efficient, it may restrict the model's ability to fully adapt to the nuances of tobacco-related content if pre-trained knowledge conflicts with domain-specific requirements. The rapidly evolving nature of social media content poses challenges for FLAASH to quickly adapt to new trends or forms of tobacco promotion without retraining. Lastly, if primarily trained in one language, the framework might need help analyzing multilingual content, which is prevalent on global social media platforms. These limitations highlight areas for future research and improvement in the FLAASH framework's approach to multi-modal tobacco content analysis.

\backmatter

\bmhead{Supplementary information} Once the ethical review is completed with privacy considerations, the Multimodal Tobacco Content Analysis Dataset (MTCAD) will be open-sourced. Given the sensitive nature of tobacco-related content and the potential inclusion of user-generated material from social media platforms, we are in the process of:
\begin{enumerate}
    \item  Ensuring full compliance with data protection regulations and platform-specific terms of service.
    \item Implementing robust anonymization techniques to protect individual privacy.
    \item Developing comprehensive guidelines for responsible use of the dataset by the research community.
\end{enumerate}

This thorough process is essential to maintain the integrity of the research and protect the privacy of individuals whose content may be included in the dataset. We anticipate completing these steps and making MTCAD publicly available in the near future. In the meantime, we are happy to provide dataset samples to reviewers upon request, subject to confidentiality agreements, to facilitate the review process while maintaining ethical standards.

\bmhead{Acknowledgements}
We would like to acknowledge Arkansas Biosciences Institute (ABI) Grant, and NSF Data Science, Data Analytics that are Robust and Trusted (DART) for their funding in supporting this research.

\section*{Declarations}

\noindent \textbf{Funding:} This work is supported by NSF DART Award \#1946391.

\noindent \textbf{Conflict of interest/Competing interests:} Not applicable

\noindent \textbf{Ethics approval and consent to participate:} Not applicable

\noindent \textbf{Consent for publication:} Not applicable

\noindent \textbf{Data availability:} Please refer to \textit{Supplementary information} for more details on data availability.  

\noindent \textbf{Materials availability:} Not applicable

\noindent \textbf{Code availability:} Implementation of the FLAASH framework will be released upon the acceptance of the article. 

\noindent \textbf{Author contribution:} Conceptualization, N.V.S.R.C., P.D.D. and K.L.; methodology, N.V.S.R.C.; programming, N.V.S.R.C.; validation, N.V.S.R.C., and K.L.; formal analysis, N.V.S.R.C.; investigation, K.L.; resources, K.L.; writing---original draft preparation, N.V.S.R.C.; writing---review and editing, P.D.D., K.L. and B.R.; visualization, N.V.S.R.C.; supervision, P.D.D., K.L. and B.R.. All authors have read and agreed to the published version of the manuscript.

% \bigskip

% Springer journals and proceedings: \url{https://www.springer.com/gp/editorial-policies}

% Nature Portfolio journals: \url{https://www.nature.com/nature-research/editorial-policies}

% \textit{Scientific Reports}: \url{https://www.nature.com/srep/journal-policies/editorial-policies}

% BMC journals: \url{https://www.biomedcentral.com/getpublished/editorial-policies}
% \end{flushleft}

\begin{appendices}

\section{Deatiled Evaluation Metrics}\label{appendix:detailed metrics}
To comprehensively assess the performance of FLAASH and compare it with existing methods, we employ a set of carefully chosen metrics. These metrics are designed to evaluate various aspects of multi-modal fusion and temporal consistency in the context of tobacco-related content analysis.

\subsection{Classification Accuracy (Cls. Acc.)}
Classification accuracy measures the model's ability to categorize tobacco-related content correctly:

\begin{equation}
    \text{Acc.} = \frac{\text{Number of correct predictions}}{\text{Total number of predictions}}
\end{equation}

This metric provides a straightforward measure of the model's overall performance in identifying and classifying tobacco-related content.

\subsection{F1 Score (F1)}
The F1 score offers a balanced measure of the model's precision and recall:

\begin{equation}
    \text{F1} = 2 \cdot \frac{\text{Precision} \cdot \text{Recall}}{\text{Precision} + \text{Recall}}
\end{equation}

Where:
\begin{equation}
    \text{Precision} = \frac{\text{True Positives}}{\text{True Positives} + \text{False Positives}}
\end{equation}
\begin{equation}
    \text{Recall} = \frac{\text{True Positives}}{\text{True Positives} + \text{False Negatives}}
\end{equation}

The F1 score is instrumental in our context, where class imbalance may exist in tobacco-related content.

\subsection{Temporal Consistency Score (TCS)}
TCS evaluates the model's ability to maintain consistent predictions across video frames:

\begin{equation}
    \text{TCS} = \frac{1}{N-1} \sum_{t=1}^{N-1} \text{sim}(\mathbf{p}_t, \mathbf{p}_{t+1})
\end{equation}

Where $N$ is the number of frames, $\mathbf{p}_t$ is the prediction for frame $t$, and $\text{sim}(\cdot,\cdot)$ is a similarity function (e.g., cosine similarity). This metric ensures that the model's predictions are stable over time, which is crucial for analyzing video content.

\subsection{Multi-modal Fusion Quality (MFQ)}
MFQ assesses how effectively the model integrates information from both visual and textual modalities:

\begin{equation}
    \text{MFQ} = \alpha \cdot \text{CCA}(\mathbf{v}, \mathbf{t}) + (1-\alpha) \cdot \text{MI}(\mathbf{f}, \mathbf{v}, \mathbf{t})
\end{equation}

Where $\mathbf{v}$ and $\mathbf{t}$ are the video and text features, respectively, $\mathbf{f}$ is the fused representation, CCA is the Canonical Correlation Analysis, MI is the Mutual Information, and $\alpha$ is a weighting factor (set to 0.5 in our experiments).

This composite metric captures the correlation between modalities (through CCA) and information preservation in the fusion process (through MI).

These metrics collectively provide a comprehensive evaluation of our FLAASH model, capturing its performance in classification, consistency over time, and effectiveness in multi-modal fusion. By consistently outperforming baseline methods across these metrics, FLAASH demonstrates its superior capability in analyzing tobacco-related content in social media videos.




\end{appendices}


\bibliography{main}% common bib file

\end{document}